\documentclass{article}
\usepackage{spconf,amsmath,graphicx}
\usepackage{bm}
\usepackage[utf8]{inputenc}
\usepackage[T1]{fontenc}
\usepackage{subfigure}
\usepackage{booktabs}
\usepackage[table]{xcolor}
\usepackage{multirow}
\usepackage{siunitx}
\usepackage{float}
\usepackage[hidelinks]{hyperref}
\title{Taking a Second Look: Correcting \\Sea Ice Forecasts with Sparse Observations}

\name{
  \begin{tabular}{c}
    Tianshuo Zhang \qquad
    Xianglei Xing$^\ast$\thanks{$^\ast$Corresponding author: xingxl@hrbeu.edu.cn.} \qquad
    Aowen Yang \\
    Jia Gao \qquad
    Wenzhe Zhai \qquad
    Shanshan Liu \qquad
    Chengtao Cai
  \end{tabular}
}

\address{
College of Intelligent Systems Science and Engineering,\\
Harbin Engineering University, Harbin 150001, China
}

\begin{document}
%
\maketitle
\begin{abstract}
Sea ice forecasts are issued several days ahead, allowing errors to accumulate while new, often sparse sea ice concentration (SIC) observations become available. We find that fixed-propagation errors concentrate near structured, high-gradient ice edges, whereas homogeneous interiors require limited propagation, suggesting that propagation distance should be state dependent. We therefore introduce \textbf{ECHO} (\textbf{E}vidence-guided \textbf{C}orrection with \textbf{H}eterogeneous pr\textbf{O}pagation), where \textbf{ECHO-Scale} adapts propagation distance while preserving correction geometry, and \textbf{ECHO-Delta} learns a bounded residual around fixed propagation. Across all 96 standard evaluation settings spanning diverse priors, observation times, sparsity levels, geometries, and noise conditions, both outperform fixed propagation. ECHO-Delta achieves the best average accuracy, while ECHO-Scale is more robust to geometry shifts. Code is available at \url{https://github.com/yingtian22/TAKING-A-SECOND-LOOK}.
\end{abstract}
\section{Introduction}
\label{sec:intro}

\begin{figure}[t]
  \centering
  \includegraphics[width=1.0\linewidth]{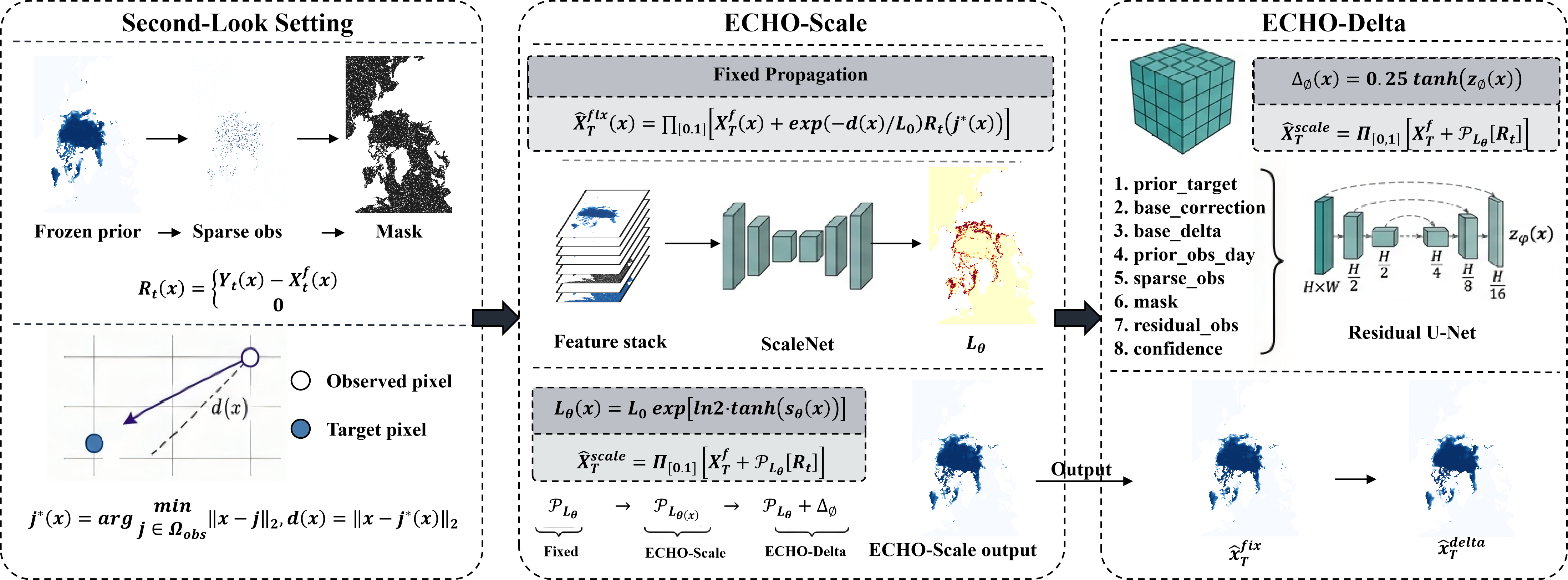}
  \caption{The Framework of ECHO.}
  \label{fig:Framework}
\end{figure}

Reliable short-term sea ice forecasts are important for safe Arctic maritime operations~\cite{wagner2020seaice,palerme2024improving}. Operational systems issue forecasts several days ahead~\cite{sakov2012topaz4}, during which forecast errors may accumulate while new satellite observations become available. We therefore consider a \emph{second-look} setting, where sparse intermediate SIC observations revise a frozen open-loop forecast before the target time. Fig.~\ref{fig:Framework} illustrates the setting and the two ECHO variants.

Sparse observations must be propagated into unobserved regions to obtain a full-field correction, a central challenge in data assimilation and sparse satellite reconstruction~\cite{carrassi2018data,cheng2023arctic,beauchamp2023fourDVarNet}. In our setting, the key question is how each observed innovation should influence its surroundings. A simple deterministic reference, \emph{Fixed Propagation}, extends the nearest-observation innovation with distance-dependent decay but applies the same rule across heterogeneous ice conditions. This leaves a central question: \emph{how far should each innovation propagate?}

Spatial error analysis shows that Fixed Propagation errors concentrate in structured, high-gradient, high-innovation ice-edge regions, while homogeneous ice and ocean interiors require little propagation. This suggests that the main limitation lies in the spatially uniform propagation extent rather than the nearest-observation geometry, motivating state-dependent adaptation of propagation distance.

Based on this diagnosis, we introduce \textbf{ECHO} (\textbf{E}vidence-guided \textbf{C}orrection with \textbf{H}eterogeneous pr\textbf{O}pagation) with two variants: \textbf{ECHO-Scale} preserves nearest-observation geometry while adapting a bounded spatially varying scale, whereas \textbf{ECHO-Delta} learns a bounded residual around Fixed Propagation. Across 96 standard settings, both improve upon Fixed Propagation and compare favorably with representative data-assimilation baselines. ECHO-Delta achieves the best average accuracy, while ECHO-Scale is lighter and more robust to geometry shifts, revealing a trade-off among accuracy, robustness, and complexity.

Our contributions are fourfold.
First, we formulate a \emph{second-look} sea ice forecast correction setting using sparse intermediate SIC observations.
Second, spatial error diagnosis identifies uniform propagation extent as the main limitation of Fixed Propagation.
Third, we introduce \textbf{ECHO}, with \textbf{ECHO-Scale} adapting propagation distance and \textbf{ECHO-Delta} adding bounded residual refinement.
Fourth, extensive evaluations demonstrate strong overall gains and an accuracy--robustness--complexity trade-off between the two variants.

\section{Related Work}
\label{sec:related}

\subsection{Sea Ice Forecasting and Data Assimilation}
Sea ice prediction spans physics-based forecasting, learning-based models, and data assimilation. TOPAZ4 couples ocean--sea ice modeling with ensemble assimilation~\cite{sakov2012topaz4}, while learning-based methods include PredRNN++~\cite{liu2021shortterm}, SICNet~\cite{ren2022sicnet}, IceDiff~\cite{xu2025icediff}, and SIFusion~\cite{xu2025sifusion}. Classical assimilation propagates innovations through prescribed statistical structure, while EnKF represents forecast uncertainty with ensembles~\cite{carrassi2018data,cheng2023arctic}; 4DVarNet addresses sparse spatiotemporal reconstruction~\cite{beauchamp2023fourDVarNet}. Unlike forecasting or reconstruction, we revise an issued target-day forecast using sparse intermediate observations.

\subsection{Learning-based Forecast Correction}
Learning-based modeling can exploit structured inductive biases~\cite{zhang2026hilnn}, while forecast post-processing reduces errors through neural calibration~\cite{rasp2018neural}, residual correction~\cite{tedesco2024bias}, and supervised refinement of operational SIC forecasts~\cite{palerme2024improving}. These methods generally learn direct mappings from available predictors to improved forecasts. ECHO instead considers a \emph{second-look} setting with a frozen forecast and sparse intermediate SIC observations. To obtain a dense target-day correction, ECHO retains a deterministic propagation reference: \textbf{ECHO-Scale} adapts its spatial influence extent, while \textbf{ECHO-Delta} learns a bounded residual around it, bridging prescribed propagation and learned correction.

\section{Methodology}
\label{sec:method}

\noindent\textbf{Second-Look Correction Formulation.}
Let $X_t^f$ and $X_T^f$ denote the frozen open-loop SIC forecasts at an
intermediate observation time $t$ and the target time $T$, respectively.
We consider $T=7$ and $t\in\{3,5\}$. Sparse SIC observations $Y_t$ are
available only over a subset of grid cells
$\Omega_{\mathrm{obs}}\subset\Omega$, where $\Omega$ denotes the valid
ocean domain. The observation innovation is
\begin{equation}
R_t(x)=
\begin{cases}
Y_t(x)-X_t^f(x), & x\in\Omega_{\mathrm{obs}},\\
0, & \text{otherwise}.
\end{cases}
\label{eq:innovation}
\end{equation}
The second-look problem is therefore to construct a dense correction
field from the sparse innovation $R_t$, while keeping the original
forecast trajectory fixed.

For every location $x$, let
\begin{equation}
j^*(x)
=
\arg\min_{j\in\Omega_{\mathrm{obs}}}\|x-j\|_2,
\qquad
d(x)=\|x-j^*(x)\|_2
\label{eq:nearest}
\end{equation}
denote the nearest observed location and its spatial distance,
respectively. We define a generic propagation operator
\begin{equation}
\mathcal{P}_{L}[R_t](x)
=
\exp\!\left(-\frac{d(x)}{L(x)}\right)
R_t\!\left(j^*(x)\right),
\label{eq:prop_operator}
\end{equation}
where $L(x)>0$ controls the propagation extent and
$\Pi_{[0,1]}(\cdot)$ denotes projection onto the valid SIC range.

\medskip
\noindent\textbf{Fixed Propagation as the Reference Operator.}
Fixed Propagation is the homogeneous case $L(x)\equiv L_0$:
\begin{equation}
\hat X_T^{\mathrm{fix}}(x)
=
\Pi_{[0,1]}
\left[
X_T^f(x)
+
\exp\!\left(-\frac{d(x)}{L_0}\right)
R_t\!\left(j^*(x)\right)
\right].
\label{eq:fixed}
\end{equation}
The validation-selected scales are $L_0=8$ for the persistence prior
and $L_0=5$ for the Direct U-Net prior. This formulation is
parameter-free at inference and preserves an explicit relationship
between observation distance and correction magnitude. Its main
restriction, however, is that the same propagation extent is imposed
throughout the spatial domain despite strongly heterogeneous local SIC
structure.

\medskip
\noindent\textbf{ECHO-Scale: Geometry-Preserving Adaptive Propagation.}
ECHO-Scale introduces the minimum state-dependent relaxation of
Eq.~\eqref{eq:fixed}: the nearest-observation assignment $j^*(x)$ and
the innovation geometry are kept unchanged, while only the propagation
extent is made spatially adaptive. Specifically, we parameterize
\begin{equation}
L_\theta(x)
=
L_0\exp\!\left[
\ln 2\,\tanh\!\left(s_\theta(x)\right)
\right]
\in[0.5L_0,2L_0].
\label{eq:scale_field}
\end{equation}
and the corrected forecast becomes
\begin{equation}
\hat X_T^{\mathrm{scale}}(x)
=
\Pi_{[0,1]}
\left[
X_T^f(x)
+
\exp\!\left(-\frac{d(x)}{L_\theta(x)}\right)
R_t\!\left(j^*(x)\right)
\right].
\label{eq:echo_scale}
\end{equation}

\begin{figure}[ht]
  \centering
  \includegraphics[width=1.0\linewidth]{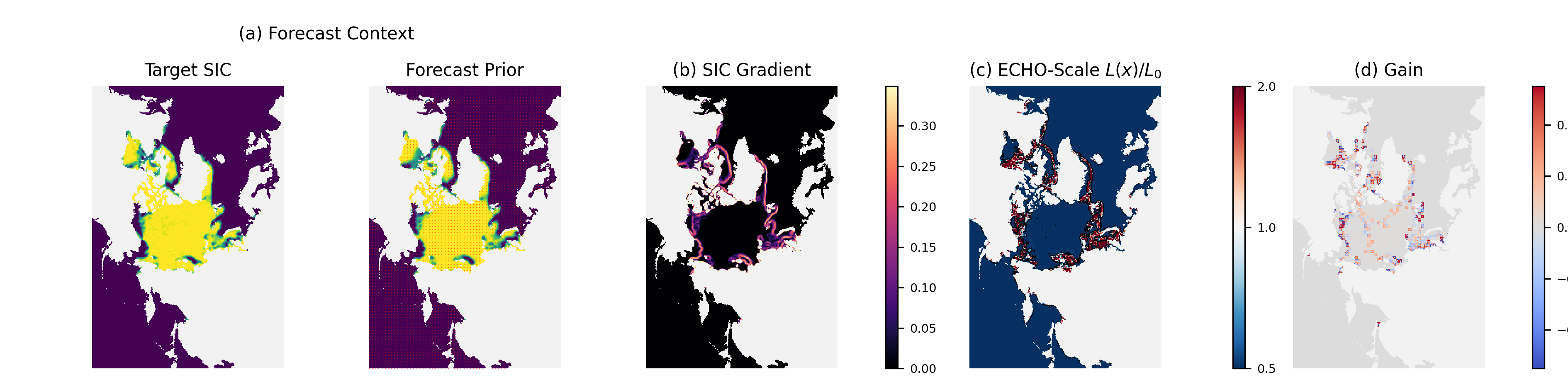}
  \caption{Spatial adaptation learned by ECHO-Scale.}
  \label{fig:echo_scale_mechanism}
\end{figure}

The scale head is zero-initialized, so $L_\theta(x)=L_0$ initially. ECHO-Scale thus starts exactly from Fixed Propagation and learns only bounded spatial deviations in propagation extent, as illustrated in Fig.~\ref{fig:echo_scale_mechanism}.

\begin{figure}[ht]
  \centering
  \includegraphics[width=1.0\linewidth]{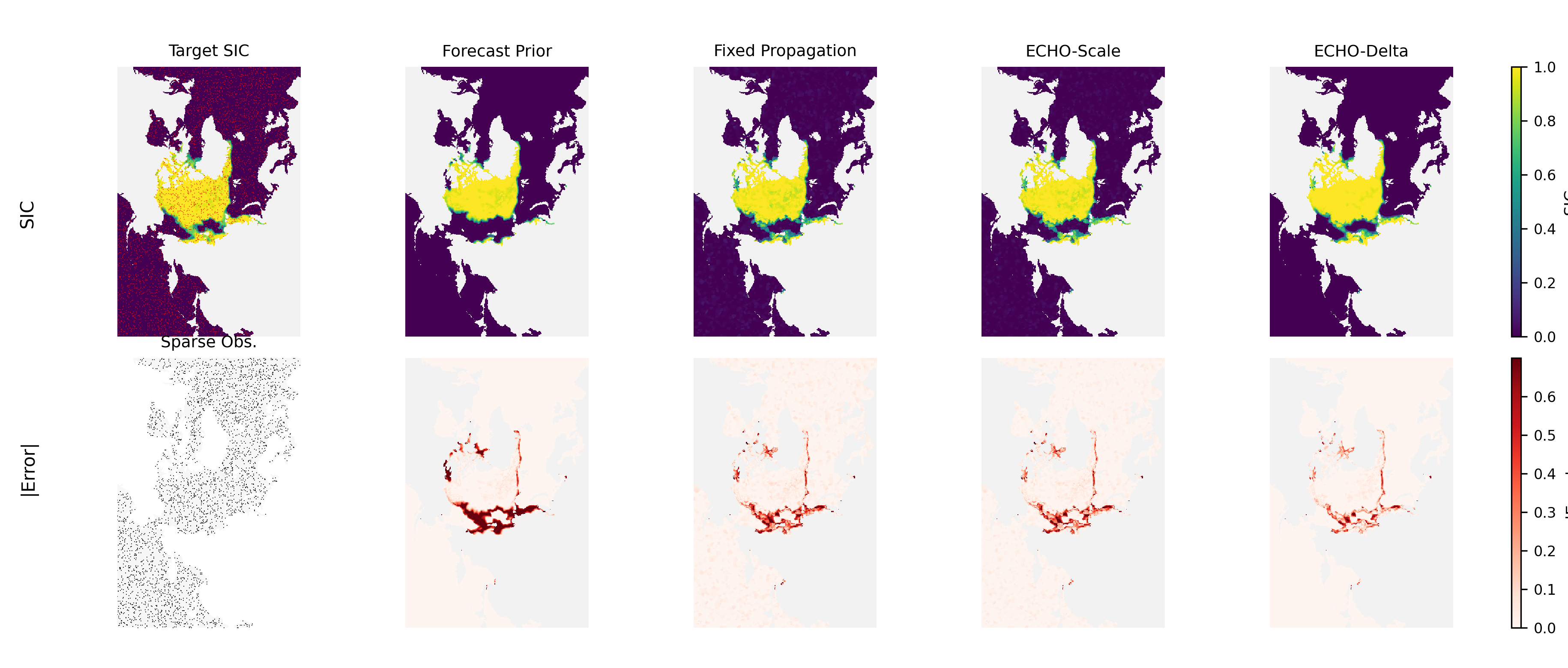}
  \caption{Qualitative comparison on a representative test case.}
  \label{fig:typical_case}
\end{figure}

\noindent\textbf{ECHO-Delta: Bounded Residual Relaxation.}
While ECHO-Scale restricts learning to the propagation scale,
ECHO-Delta introduces a second, more expressive relaxation around the
same deterministic reference. Let
\begin{equation}
\Delta_\phi(x)
=
\delta_{\max}
\tanh\!\left(z_\phi(x)\right),
\qquad
\delta_{\max}=0.25,
\label{eq:echo_delta_residual}
\end{equation}
denote a bounded residual correction. The final prediction is
\begin{equation}
\hat X_T^{\mathrm{delta}}(x)
=
\Pi_{[0,1]}
\left[
\hat X_T^{\mathrm{fix}}(x)
+
\Delta_\phi(x)
\right].
\label{eq:echo_delta}
\end{equation}

The bounded parameterization prevents unrestricted departures from the
reference correction while allowing errors not captured by an
isotropic distance-decay model to be compensated directly.

\begin{figure}[ht]
  \centering
  \includegraphics[width=1.0\linewidth]{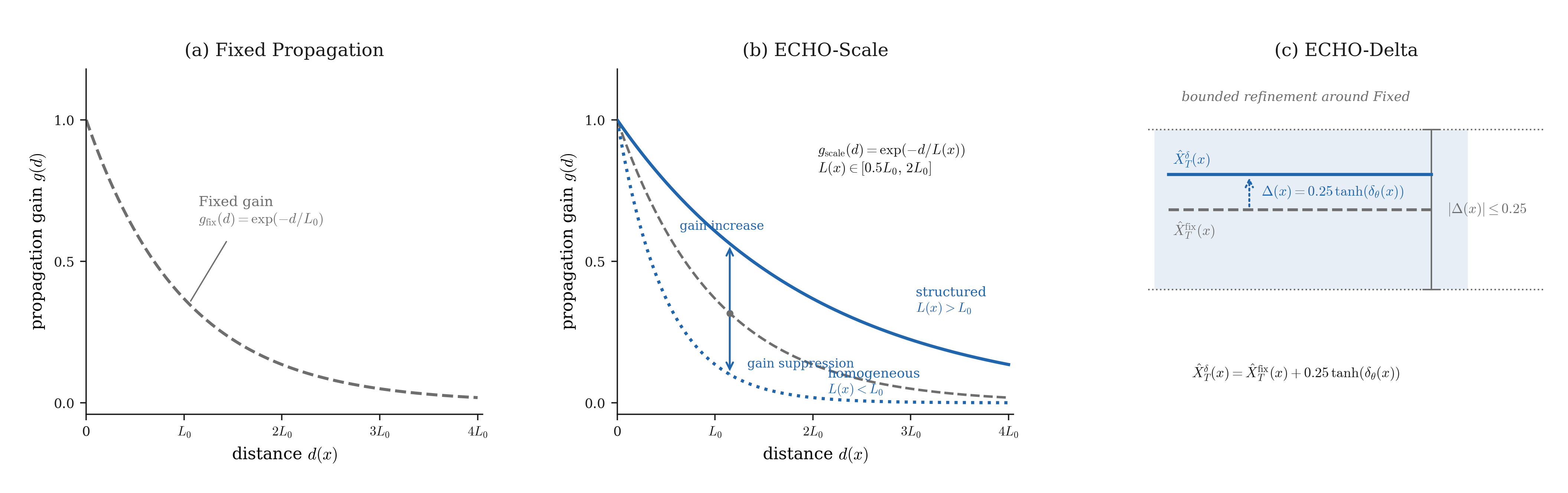}
  \caption{Comparison of Fixed Propagation and ECHO variants.}
  \label{fig:echo_mechanism_comparison}
\end{figure}

The two variants therefore instantiate two levels of relaxation of the
same reference operator:
\begin{equation}
\underbrace{\mathcal{P}_{L_0}}_{\text{Fixed}}
\;\longrightarrow\;
\underbrace{\mathcal{P}_{L_\theta(x)}}_{\text{ECHO-Scale}}
\;\longrightarrow\;
\underbrace{\mathcal{P}_{L_0}+\Delta_\phi}_{\text{ECHO-Delta}},
\label{eq:echo_hierarchy}
\end{equation}
where ECHO-Scale adapts only the influence extent, whereas ECHO-Delta adds bounded residual refinement. This hierarchy tests whether scale adaptation alone suffices or residual correction is required, as illustrated in Fig.~\ref{fig:echo_mechanism_comparison}.

\noindent\textbf{Learning Objective.}
Let $e(x)=\hat X_T(x)-X_T(x)$ denote the prediction error over valid
ocean pixels. Both variants minimize
\begin{equation}
\mathcal{L}
=
\frac{1}{|\Omega_{\mathrm{valid}}|}
\sum_{x\in\Omega_{\mathrm{valid}}}
\rho(e(x)),
\label{eq:training_general}
\end{equation}
where the error penalty is chosen according to the degree of model
flexibility:
\begin{equation}
\rho(e)
=
\begin{cases}
e^2,
& \text{ECHO-Scale},\\[2mm]
|e|+\lambda e^2,
& \text{ECHO-Delta},
\end{cases}
\qquad
\lambda=0.2.
\label{eq:training}
\end{equation}
ECHO-Scale uses the quadratic loss, whereas ECHO-Delta uses the mixed
$\ell_1$--$\ell_2$ loss; predictions are projected onto $[0,1]$ before
loss evaluation.

\section{Experiments}

\noindent\textbf{Experimental Setup.}
We use the NOAA/NSIDC Sea Ice Concentration CDR (G02202 v5)~\cite{meier2024g02202} from 2014--2020 on a $448\times304$ polar-stereographic grid, with 1819/365/360 train/validation/test samples. We evaluate lead day 7 using persistence and a pretrained Direct U-Net as frozen priors. Sparse observations at $t\in\{3,5\}$ use $\{5\%,10\%,30\%\}$ coverage, four geometries (random, edge-biased, stripe, coarse-grid), and noise $\sigma\in\{0,0.03\}$, yielding 96 settings with 360 test cases each. Metrics include RMSE, MAE, ice-edge RMSE, MIZ RMSE, and sea-ice extent error. Baselines include residual interpolation, nudging, localized OI, and EnKF-based assimilation; OI uses validation-selected $L=2$ and $r_{\mathrm{loc}}=4$. Both ECHO variants use AdamW with learning rate $5\times10^{-4}$, batch size 4, 20 epochs, patience 6, and three seeds.

\begin{figure}[ht]
  \centering
  \includegraphics[width=1.0\linewidth]{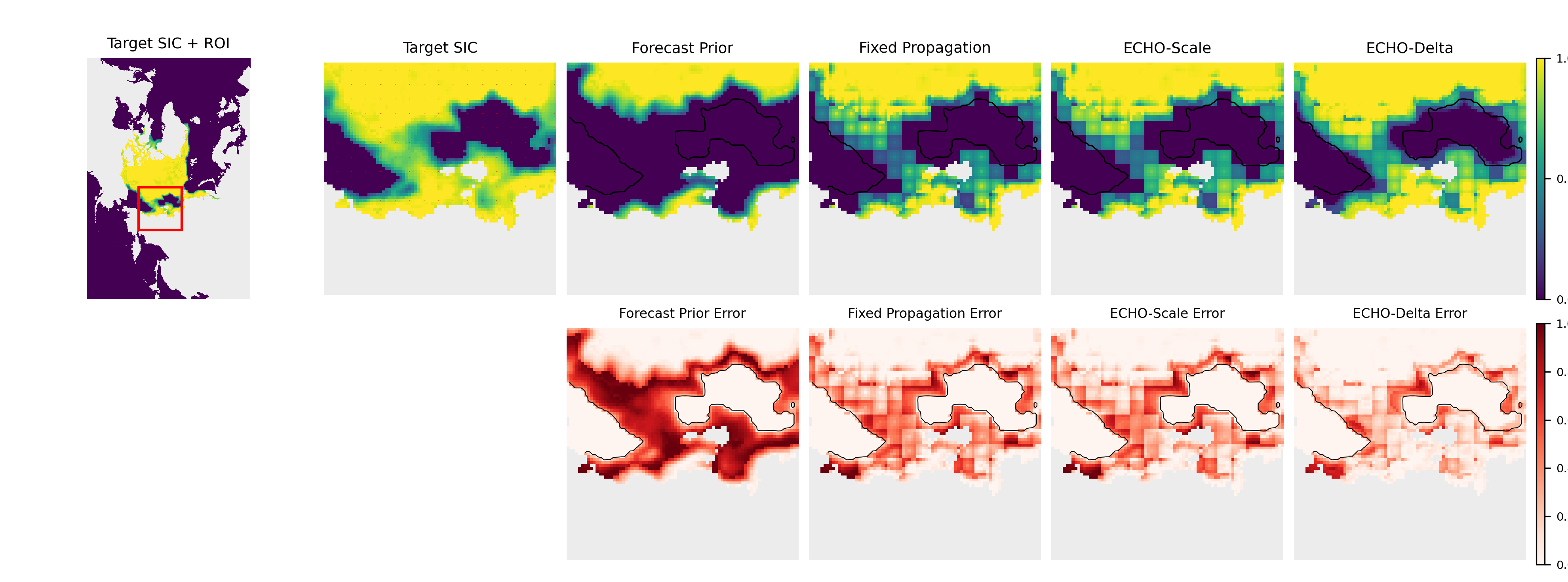}
  \caption{Qualitative comparison at the ice edge.}
  \label{fig:ice_edge_zoom}
\end{figure}

\noindent\textbf{Main Results.}
Table~\ref{tab:main_results} summarizes performance over all 96 settings, with ECHO reported as mean$\pm$std over three training seeds and the remaining baselines using their frozen formal results. ECHO-Delta achieves the lowest average RMSE ($0.06249\pm0.00028$), followed by ECHO-Scale ($0.06735\pm0.00003$). Using the seed-42 paired protocol, both variants outperform Fixed Propagation in all 96 settings, with mean $\Delta$RMSEs of $-0.001634$ for ECHO-Scale (95\% CI $[-0.001745,-0.001523]$) and $-0.006765$ for ECHO-Delta (95\% CI $[-0.007356,-0.006177]$). Against validation-tuned Localized OI, ECHO-Scale is better on average despite winning 61 of 96 settings ($\Delta=-0.000917$, 95\% CI $[-0.001333,-0.000507]$), while ECHO-Delta wins all 96 settings ($\Delta=-0.006048$, 95\% CI $[-0.006777,-0.005338]$). Fig. ~\ref{fig:ice_edge_zoom} highlights the improvement around the ice edge, while Fig.~\ref{fig:typical_case} provides a full-domain qualitative comparison.

\begin{table}[ht]
  \centering
  \caption{Results over 96 settings. ECHO reports 3-seed means; CIs use seed 42.}
  \label{tab:main_results}

  \fontsize{9}{10.5}\selectfont
  \setlength{\tabcolsep}{2.2pt}

  \begin{tabular}{@{}lcccc@{}}
    \toprule
    \bfseries Method &
    \bfseries RMSE $\downarrow$ &
    \bfseries MAE $\downarrow$ &
    \bfseries \shortstack{Edge\\RMSE $\downarrow$} &
    \bfseries \shortstack{MIZ\\RMSE $\downarrow$} \\
    \midrule

    Prior
    & 0.08161 & 0.01753 & 0.15568 & 0.28302 \\

    Residual Interp.
    & 0.07172 & 0.02065 & 0.13921 & 0.24391 \\

    Fixed Prop.
    & 0.06897 & 0.01850 & 0.13463 & 0.24043 \\

    Nudging~\cite{carrassi2018data}
    & 0.06950 & 0.01911 & 0.13499 & 0.23936 \\

    Localized OI~\cite{shao2023arcticoi}
    & 0.06825 & 0.01794 & 0.13351 & 0.23841 \\

    EnKF-PertObs~\cite{cheng2023arctic}
    & 0.07989 & 0.01934 & 0.15312 & 0.27226 \\

    4DVarNet~\cite{beauchamp2023fourDVarNet}
    & 0.06909 & 0.02645 & 0.13581 & \textbf{0.22662} \\

    \rowcolor{gray!20}
    \textbf{ECHO-Scale}
    & \underline{0.06735}
    & \underline{0.01736}
    & \underline{0.13209}
    & 0.23732 \\

    \rowcolor{gray!20}
    \textbf{ECHO-Delta}
    & \textbf{0.06249}
    & \textbf{0.01309}
    & \textbf{0.12477}
    & \underline{0.23085} \\
    \bottomrule
  \end{tabular}
\end{table}

We further compare ECHO with end-to-end forecasting models using sparse observations. Since these models do not use the frozen target-day prior $X_T^f$, they are complementary rather than matched-protocol baselines. Only IceDiff-FM+obs improves upon the original prior on average, yet remains worse than Fixed Propagation and both ECHO variants.

\begin{table}[ht]
  \centering
  \caption{End-to-end comparison with sparse observations over 96 settings.}
  \label{tab:e2e_obs}

  \fontsize{9}{10.5}\selectfont
  \setlength{\tabcolsep}{1.2pt}

  \begin{tabular}{@{}lccccc@{}}
    \toprule
    \bfseries Method &
    \bfseries RMSE $\downarrow$ &
    \bfseries MAE $\downarrow$ &
    \bfseries \shortstack{Edge\\RMSE $\downarrow$} &
    \bfseries \shortstack{MIZ\\RMSE $\downarrow$} &
    \bfseries \shortstack{Uses\\$X_T^f$} \\
    \midrule

    Original Prior
    & 0.08161
    & 0.01753
    & 0.15568
    & 0.28302
    & -- \\

    IceDiff-FM~\cite{xu2025icediff}+obs
    & 0.07729
    & 0.02647
    & 0.14903
    & 0.25535
    & no \\

    Unicorn~\cite{park2025unicorn}+obs
    & 0.08307
    & 0.02872
    & 0.15704
    & 0.24721
    & no \\

    ConvLSTM~\cite{shi2015convlstm}+obs
    & 0.08736
    & 0.02669
    & 0.16397
    & 0.28887
    & no \\

    SICNet-TSAM~\cite{ren2022sicnet}+obs
    & 0.11716
    & 0.05096
    & 0.20206
    & 0.29682
    & no \\

    Fixed Prop.
    & 0.06897
    & 0.01850
    & 0.13463
    & 0.24043
    & yes \\

    \rowcolor{gray!20}
    \textbf{ECHO-Scale}
    & \underline{0.06733}
    & \underline{0.01735}
    & \underline{0.13206}
    & \underline{0.23701}
    & yes \\

    \rowcolor{gray!20}
    \textbf{ECHO-Delta}
    & \textbf{0.06220}
    & \textbf{0.01295}
    & \textbf{0.12432}
    & \textbf{0.23202}
    & yes \\

    \bottomrule
  \end{tabular}
\end{table}

\noindent\textbf{Global Retuning and Spatial Scale Adaptation.}
To test whether ECHO-Scale simply compensates for a suboptimal global scale, we re-optimize $L$ on the validation set. The optimum remains $L=8$ for persistence and shifts from $5$ to $3.75$ for Direct U-Net. This retuning improves Fixed Propagation RMSE only from 0.068966 to 0.068936, explaining just 1.8\% of the gap to ECHO-Scale. ECHO-Scale still wins all 96 settings (mean $\Delta$RMSE $=-0.001604$, 95\% CI $[-0.001719,-0.001488]$). Independent scale sweeps further show that higher-gradient regions favor larger propagation extents, consistent with the learned spatial scales.


\begin{table}[ht]
  \centering
  \caption{Validation results by prior-SIC gradient regime.}
  \label{tab:scale_analysis}

  \fontsize{9}{10.5}\selectfont
  \setlength{\tabcolsep}{2.0pt}

  \begin{tabular}{@{}lcccc@{}}
    \toprule
    \bfseries Regime &
    \bfseries \shortstack{Fixed\\RMSE $\downarrow$} &
    \bfseries \shortstack{ECHO-Scale\\RMSE $\downarrow$} &
    \bfseries \shortstack{Best\\$\bm{L/L_0}$} &
    \bfseries \shortstack{Learned\\$\bm{L(x)/L_0}$} \\
    \midrule

    Low (25\%)
    & 0.0151
    & \textbf{0.0135}
    & 0.25
    & 0.51 \\

    Medium (47\%)
    & 0.0393
    & \textbf{0.0376}
    & 0.50
    & 0.54 \\

    High (28\%)
    & 0.1185
    & \textbf{0.1160}
    & 1.25
    & 0.73 \\

    \bottomrule
  \end{tabular}
\end{table}

\noindent\textbf{Sensitivity to Observation Sparsity.}
Performance improves as observation coverage increases from 5\% to 30\%. Even at 5\% coverage, both ECHO variants outperform Fixed Propagation in all 32 matched seed-42 settings. ECHO-Scale benefits most from sparse observations, while ECHO-Delta maintains larger gains across all coverage levels.


\begin{table}[ht]
  \centering
  \caption{RMSE under different observation coverages.}
  \label{tab:sparsity}

  \fontsize{9}{10.5}\selectfont
  \setlength{\tabcolsep}{3pt}

  \begin{tabular}{@{}lccc@{}}
    \toprule
    \bfseries Coverage &
    \bfseries Fixed $\downarrow$ &
    \bfseries ECHO-Scale $\downarrow$ &
    \bfseries ECHO-Delta $\downarrow$ \\
    \midrule

    5\%
    & 0.07049
    & \underline{0.06862}
    & \textbf{0.06400} \\

    10\%
    & 0.06890
    & \underline{0.06720}
    & \textbf{0.06226} \\

    30\%
    & 0.06751
    & \underline{0.06623}
    & \textbf{0.06120} \\

    \bottomrule
  \end{tabular}
\end{table}

\noindent\textbf{Robustness and Generalization.}
Across three mask and noise realizations, both ECHO variants outperform Fixed Propagation in all 96 settings with negligible RMSE variation; date-matched filtering preserves the paired rankings. Under sensor-inspired geometries, ECHO-Scale is never worse than Fixed (88 wins, 8 ties), while ECHO-Delta retains the lowest average RMSE but shows 36 reversals, all under the Direct U-Net prior, and wins all 48 persistence settings. Localized OI performs best with dense, locally distributed observations but weakens under stripe sampling and at 5\% coverage. Overall, ECHO-Delta favors average accuracy, whereas ECHO-Scale provides stronger geometry-shift consistency.

\noindent\textbf{Ablation Study.}
Table~\ref{tab:ablation} ablates the input guidance of ECHO-Delta.
Removing either the distance-decay confidence or the Fixed increment
degrades performance, while jointly removing the Fixed-guidance features
causes the largest RMSE increase. The confidence ablation degrades most
under stripe masks, where observations are spatially clustered.






\begin{table}[ht]
  \centering
  \caption{ECHO-Delta ablation study with training seed 42.}
  \label{tab:ablation}

  \fontsize{9}{10.5}\selectfont
  \setlength{\tabcolsep}{2.0pt}

  \begin{tabular}{@{}lcccc@{}}
    \toprule
    \bfseries Variant &
    \bfseries RMSE $\downarrow$ &
    \bfseries $\Delta$RMSE &
    \bfseries \shortstack{Edge\\RMSE $\downarrow$} &
    \bfseries \shortstack{MIZ\\RMSE $\downarrow$} \\
    \midrule

    \rowcolor{gray!20}
    \textbf{Full ECHO-Delta}
    & \textbf{0.06220}
    & --
    & \textbf{0.12432}
    & \textbf{0.23202} \\

    w/o Confidence
    & 0.06294
    & +0.00073
    & 0.12606
    & 0.23679 \\

    w/o Base-Delta
    & 0.06290
    & +0.00069
    & 0.12583
    & 0.23424 \\

    w/o Fixed-Guidance
    & 0.06315
    & +0.00095
    & 0.12644
    & 0.23725 \\

    \bottomrule
  \end{tabular}
\end{table}

\noindent\textbf{Efficiency and Complexity.}
ECHO-Scale uses 76.9K parameters and 136 MiB peak GPU memory, versus 7.85M and 397 MiB for ECHO-Delta. Mean latency is 14.43, 42.24, and 39.42 ms/case for Fixed, ECHO-Scale, and ECHO-Delta, respectively, making ECHO-Scale substantially lighter while retaining comparable learned-model latency.

\section{Conclusion}
\label{sec:conclusion}

We studied second-look sea ice forecast correction from sparse observations and identified uniform propagation extent as a key limitation of Fixed Propagation. ECHO-Scale adapts this extent, while ECHO-Delta adds bounded residual refinement. Both improve upon Fixed Propagation across all 96 settings; ECHO-Delta achieves the best average accuracy, whereas ECHO-Scale offers stronger geometry-shift consistency at lower complexity.

\bibliographystyle{IEEEbib}
\bibliography{refs}

\section{Acknowledgment}
This work was supported by the National Natural Science Foundation of China under Grant No. 62676111, and the HEU CISSE Decanal Innovation Fund for Ph.D. Students. The authors have no relevant financial or nonfinancial interests to disclose.

\section{Compliance with Ethical Standards}
This study uses publicly available sea ice data and involves no human or animal subjects; ethical approval was not required.

\end{document}